\documentclass[conference]{IEEEtran}
\IEEEoverridecommandlockouts
\renewcommand\IEEEkeywordsname{Keywords}
\usepackage{cite}
\usepackage{amsmath,amssymb,amsfonts}
\usepackage{algorithmic}
\usepackage{graphicx}
\graphicspath{{images/}}
\usepackage{textcomp}
\usepackage{xcolor}
\usepackage[caption=false,font=scriptsize,labelfont=sf,textfont=sf]{subfig}
\usepackage[hidelinks]{hyperref}
\def\BibTeX{{\rm B\kern-.05em{\sc i\kern-.025em b}\kern-.08em
    T\kern-.1667em\lower.7ex\hbox{E}\kern-.125emX}}

\makeatletter

\begin{document}

\title{Cross-domain Sentiment Classification in Spanish}

% \author{\IEEEauthorblockN{Double-Blind Revision}}
\author{\IEEEauthorblockN{Lautaro Estienne}
\IEEEauthorblockA{\textit{Facultad de Ingeniería} \\
\textit{Universidad de Buenos Aires (UBA)}\\
Buenos Aires, Argentina \\
lestienne@fi.uba.ar}
\and
\IEEEauthorblockN{Matias Vera}
\IEEEauthorblockA{\textit{CSC-CONICET} \\
\textit{and Facultad de Ingeniería UBA}\\
Buenos Aires, Argentina  \\
mvera@fi.uba.ar}
\and
\IEEEauthorblockN{Leonardo Rey Vega}
\IEEEauthorblockA{\textit{CSC-CONICET} \\
\textit{and Facultad de Ingeniería UBA}\\
Buenos Aires, Argentina  \\
lrey@fi.uba.ar}
}
% \and
% \and
% \and

\maketitle

\begin{abstract}
% This document is a model and instructions for \LaTeX.
% This and the IEEEtran.cls file define the components of your paper [title, text, heads, etc.]. *CRITICAL: Do Not Use Symbols, Special Characters, Footnotes, 
% or Math in Paper Title or Abstract.
Sentiment Classification is a fundamental task in the field of Natural Language Processing, and has very important academic and commercial applications. It aims to automatically predict the degree of sentiment present in a text that contains opinions and subjectivity at some level, like product and movie reviews, or tweets. This can be really difficult to accomplish, in part, because different domains of text contains different words and expressions. In addition, this difficulty increases when text is written in a non-English language due to the lack of databases and resources. As a consequence, several cross-domain and cross-language techniques are often applied to this task in order to improve the results. In this work we perform a study on the ability of a classification system trained with a large database of product reviews to generalize to different Spanish domains. Reviews were collected from the MercadoLibre website from seven Latin American countries, allowing the creation of a large and balanced dataset. Results suggest that generalization across domains is feasible though very challenging when trained with these product reviews, and can be improved by pre-training and fine-tuning the classification model.

\end{abstract}

\section{Introduction}\label{sec:intro}
With the growing availability and popularity of opinion-rich resources such as social networks, e-commerce platforms and blogs, there has been an interest to correctly understanding what people think. For this reason, the fields of Opinion Mining and Sentiment Analysis (SA), which deal with the computational treatment of opinion, sentiment, and subjectivity in text, have lately become hot topics in research and industrial applications~\cite{panglee}\cite{Birjali2021}.

As part of the SA field, Sentiment Classification (SC) is the task of automatically predicting the sentiment level of a text. That is, to assign a value (usually an integer from 1 to 5) to a piece of text that represents the sentiment information present in it (usually, 1 being very negative and 5 being very positive). In order to perform this prediction, a classification model can be trained using data from the Internet, like product or movie reviews and texts extracted from Social Networks like Twitter. Each of these sources are known in the literature as the text domain. 

Due to the massive access to data, opinions found in the Internet can span so many different domains  where annotated training data for all of them~\cite{peng2018} is not always available. For example, product reviews can be much more easy to collect that annotated tweets, because reviews are usually labeled with the user's star rating, while tweets need to be manually labeled. This has motivated much research on cross-domain sentiment classification (CDSC) which aims to transfer the knowledge from label rich domain (source domain) to the label few domain (target domain)~\cite{adversarial}.

\subsection{The Cross-Domain Approach}

In recent years, the raise to prominence of Deep Learning has been essential to perform knowledge transfer from one task to another. Specifically, pivot-based methods~\cite{blitzer2007}, domain-invariant features extraction~\cite{peng2018} and multitask learning~\cite{ijcai2019} are some of the techniques found in the literature to improve the performance of CDSC systems. All of these methods need to train a dedicated NLP model from scratch for every new domain with its own specialized training data. Alternatively, substantial work has shown that unsupervised pre-trained language models on large text corpus are beneficial for text classification and other NLP tasks, which can avoid training a new model from scratch~\cite{google2021}. Pre-training a deep model consists in learning some or all of its multiple layers using a large amount of easy-to-collect data on a general task like Language Modeling. Then, these layers are used to initialize the model that performs a specific task like sentiment classification.
Recent advances in developing pre-trained models like BERT~\cite{bert} and GPT~\cite{gpt2} through language modeling have proved that transfer learning can improve performance in several supervised task by fine-tuning the model with supervised data. Moreover, most recent models like GPT-3~\cite{gpt3} achieved promising results in some tasks without the need of fine-tuning, i.e., with a zero-shot configuration. The term ``zero-shot'' is used to denote that, once the model has been pre-trained, it is evaluated on additional data without any additional training effort.

In this work we address the task of Spanish CDSC by training a model on product reviews from the MercadoLibre website and evaluating it on three others datasets: the Multilingual Amazon Reviews Corpus (the Spanish portion)~\cite{amazon}, the MuchoCine Reviews Corpus~\cite{muchocine} and the TASS General Coprus (2012)~\cite{tass2012}, which are described in Section~\ref{sec:data}. We perform experiments using two different models, one trained from scratch and one pre-trained on a Language Model task. Additionally, we evaluate these models in each of the above mentioned datasets in two different configurations:
\begin{itemize}
    \item \textit{fine-tuning}, where a model is pre-trained to perform SC in the context of products reviews and then is trained again with the target dataset to perform SC in a new domain.
    \item \textit{zero-shot}, where a model is trained to perform SC in the context of product reviews and evaluated in the target domain without any additional training.
\end{itemize}

\subsection{Related Works}

Despite the interest in CDSC research, few works have been found in the literature about cross-domain in Spanish sentiment classification. A cross-domain study was performed in \cite{gonazalez2014}, where the authors uses different lexicons to identify the adaptability of some words in different domains. Other than that, some works~\cite{brooke2009}~\cite{Miranda2016} perform a cross-linguistic approach in which some of the cross-domain techniques used in English are applied to Spanish text. To the best of our knowledge, this is the first work where a cross-domain classification in Spanish is made using pre-trained transformers model. Additionally, no other work where a Spanish model with a zero-shot configuration was evaluated was found in the literature.

The rest of the paper is organized as follows. Section \ref{sec:data} describe the MeLiSA (MercadoLibre for Sentiment Analysis) dataset, which was created using the MercadoLibre API\footnote{https://developers.mercadolibre.com.ar} and consists on a large collection of product reviews from seven Latin American countries. This Section also describes the additional datasets used for the cross-domain study. Section \ref{sec:models} describe the two models used to perform sentiment classification, and how this classification was implemented and evaluated. In Section \ref{sec:results_discuss} the results of the cross-domain sentiment classification are shown and analyzed. Finally, Section \ref{sec:conclusions} contains the main conclusions from the study and possible future work.

\section{MeLiSA: Mercado Libre for Sentiment Analysis}\label{sec:data}
The idea of building a database from product reviews was inspired in~\cite{panglee}, which contains a very comprehensive analysis of opinion mining on this type of websites. Many of the criteria on how to filter reviews and balance the amount of review in each class were also drawn from that work. Other useful sources were~\cite{socher2013} and~\cite{maas2011}, which describe similar English databases, and~\cite{navasloro2020}, which is a review on Spanish SA databases.

\subsection{Data Preparation}

\begin{figure}[t]
    \centering
    \includegraphics[width=.5\textwidth]{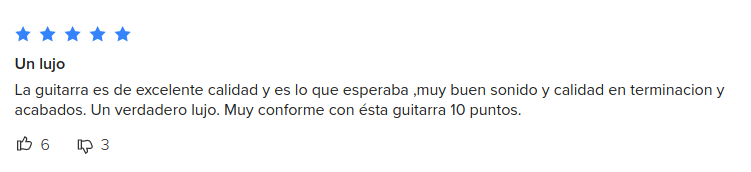}
    \caption{Sample of a product review taken from the MercadoLibre website.}
    \label{fig:MLreview}
\end{figure}

Figure \ref{fig:MLreview} shows a typical review from the MercadoLibre website. Each review consists mainly on a \textit{body comment}, a \textit{title} and a \textit{number of stars}, which is considered as the type of sentiment present in the body, i.e., from ``very negative'' (one star) through ``very positive'' (five stars). In addition to this features, each review has the number of likes and dislikes given to the comment by other users. The difference between the number of likes and dislikes is called \textit{valorization} and it could be used as a metric of how useful was the comment to other users. Comments with high valorization were prioritized at the time of selecting the reviews.

The MeLiSA database was build using the MercadoLibre API to access to the data. This API does not require any kind of authentication and it is open source, so all data in the database was public at the time it was downloaded. All reviews were downloaded between August 2020 and January 2021. Because MercadoLibre commercial presence is not the uniform through Latin America, only the countries with more reviews were considered for the their inclusion in the database. These countries were Argentina, Colombia, Peru, Uruguay, Chile, Venezuela and Mexico. In addition, in order to avoid any correlation between the country from which the review belong and the star rate we select a balanced amount of reviews per star rate inside the same country.

Each product published in MercadoLibre belongs to a specific category. Because it is likely that reviews of products from the same category have similar vocabulary, correlation between the product category and the star rate can happen if the data is not correctly balanced. As an instance, if category A has a tendency to contains low star rate reviews, models trained with that kind of data will tend to assign low star rate to reviews that present words from A's semantic field. In order to avoid that kind of bias, we manually clustered all the MercadoLibre product categories, resulting in a big five group: ``Food and drinks'', ``Arts and entertainment'', ``Home'', ``Personal Health'' and ``Electronics and Technology''. We then selected the same amount of reviews for each category and star rate.

With this approach we were able to select $487,\!368$ Spanish reviews from the MercadoLibre website. These samples were used to create a train, validation and test partition of the dataset with the following criteria: each split should be balanced by star rate and the test set should have the same amount of samples for the same category, country and star rate. The number of samples per split is shown in Table \ref{tab:data_splits}.

\begin{table}[ht!]
    \centering
    \caption{Number of data samples per split}
    \label{tab:data_splits}
    \begin{tabular}{r|ccc}
         & $|$Train$|$ & $|$Val$|$ & $|$Test$|$ \\\hline
         MeLiSA & $439,\!250$ & $23,\!118$ & $25,\!000$ \\
         Amazon & $200,\!000$ & $5,\!000$ & $5,\!000$ \\
         MuchoCine & $3,\!290$ & $193$ & $388$ \\
         TASS (2012) & $5,\!192$ & $574$ & $39,\!382$ \\
    \end{tabular}
\end{table}

\begin{figure*}[!t]
    \centering
    \subfloat[]{\includegraphics[width=.40\textwidth]{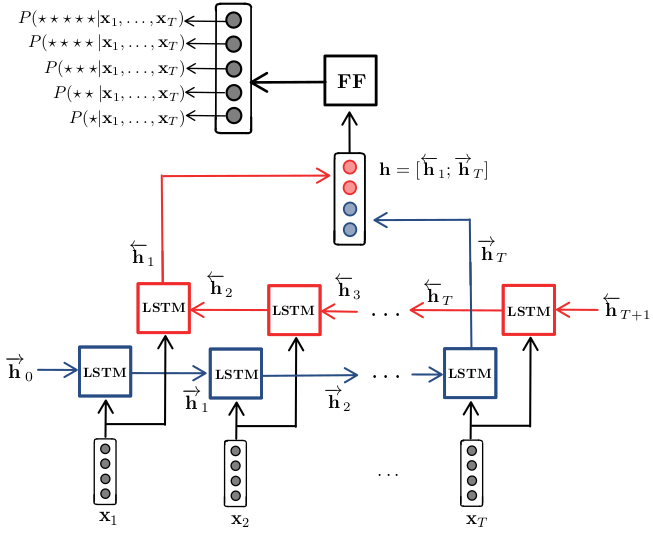}
    \label{fig:blstm_classifier}}
    \hfil
    \subfloat[]{\includegraphics[width=.58\textwidth]{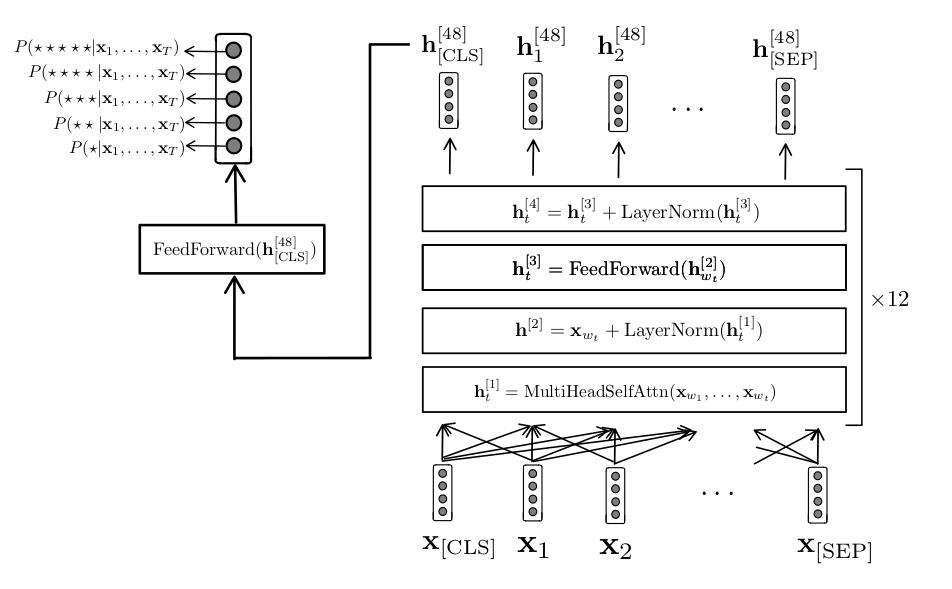}
    \label{fig:bert_classifier}}
    \caption{Classification models. (a) BiLSTM. (b) BERT.}
    \label{fig:classification_models}
\end{figure*}

\subsection{Additional Databases}

In addition to MeLiSA, Table \ref{tab:data_splits} shows the number of samples in every split of three others datasets. These datasets were used to evaluate the task of sentiment classification when trained with MeLiSA as the source domain. We briefly describe these databases below:

\begin{itemize}
    \item \textbf{Amazon}~\cite{amazon}. This database is a collection of reviews on products sold on Amazon\footnote{https://www.amazon.com/}, collected between 2015 and 2019. It was created with the purpose of studying multilingual language understanding and contains comments in English, Japanese, German, French, Spanish and Chinese. For this work, the Spanish portion of the database was used, which consists of a training portion (200,000 reviews), a validation portion (5,000 reviews), and a test portion (5,000 reviews).
    
    \item \textbf{MuchoCine}~\cite{muchocine}. This database consists in 3,871 movie reviews extracted from the MuchoCine\footnote{http://www.muchocine.net/} website. It contains comments with their respective titles and a star rate between 1 and 5, given by the movie reviewer. This database is open source and relatively balanced by star rate, but does not have a standard division into training, validation or testing, and simply consists on a single set of review-rate sample pairs. For this work, a random partition was used in order to reserve 85\% of the samples for training, 5\% for validation and 10\% for testing. This partition remained the same for all experiments.
    
    \item \textbf{TASS (2012)}~\cite{tass2012}. The SEPLN Semantic Analysis Workshop (\textit{Taller de Análsis de sentimientos de la SEPLN}, or TASS) is a workshop that has been part of the Spanish Natural Language Processing Society (SEPLN) since 2012. Since that year, a series of databases have been developed to facilitate the study of different NLP tasks. The general corpus, created for its first edition in 2012, consists of a series of semi-automatically tagged Spanish tweets with 6 possible categories that denote the degree of sentiment of the tweet: P+ (very positive), P (positive), NEU (neutral), N (negative), N+ (very negative) and NONE (no sentiment). The corpus version is available for download after registration and contains 7,219 training tweets and 60,798 test tweets. It is not balanced by star rate. In order to compare the results with the rest of the databases used in this work, both sets were modified so that only those tweets labeled with P+, P, NEU, N or N+ (that is, without NONE) are included.
\end{itemize}

\section{Cross-Domain Sentiment Classification}\label{sec:models}

\begin{figure*}
    \centering
    \subfloat[]{\includegraphics[width=.33\textwidth]{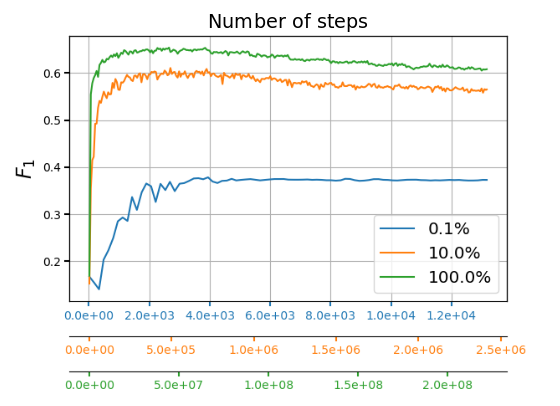}\label{fig:melisa_pretrain}}
    \hfil
    \subfloat[]{\includegraphics[width=.33\textwidth]{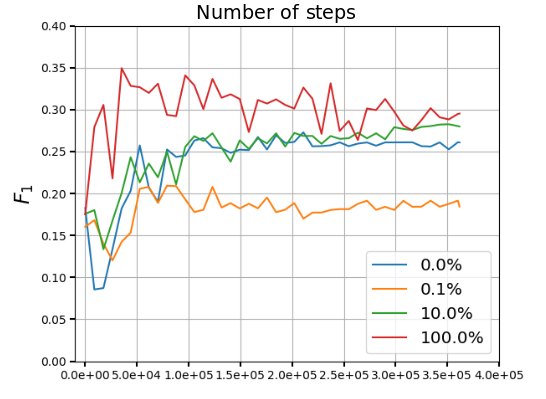}\label{fig:finetune_cine}}
    \hfil
    \subfloat[]{\includegraphics[width=.33\textwidth]{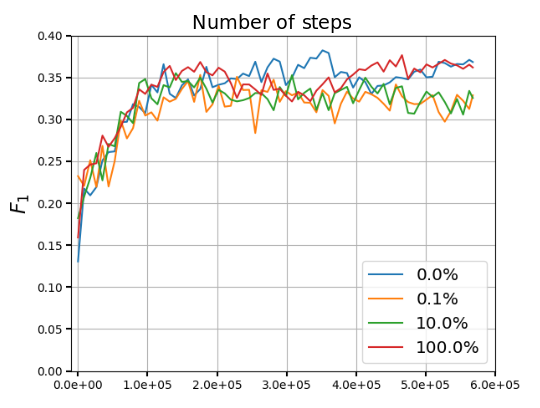}\label{fig:finetune_tass}}
    
    \subfloat[]{\includegraphics[width=.33\textwidth]{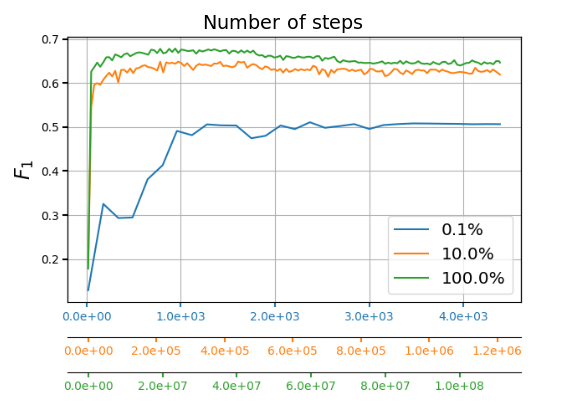}\label{fig:melisa_pretrain_bert}}
    \hfil
    \subfloat[]{\includegraphics[width=.33\textwidth]{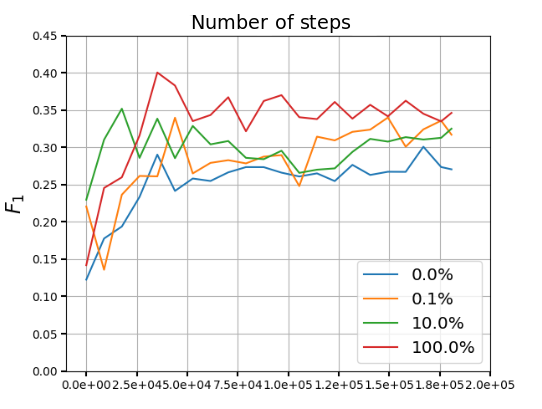}\label{fig:finetune_cine_bert}}
    \hfil
    \subfloat[]{\includegraphics[width=.33\textwidth]{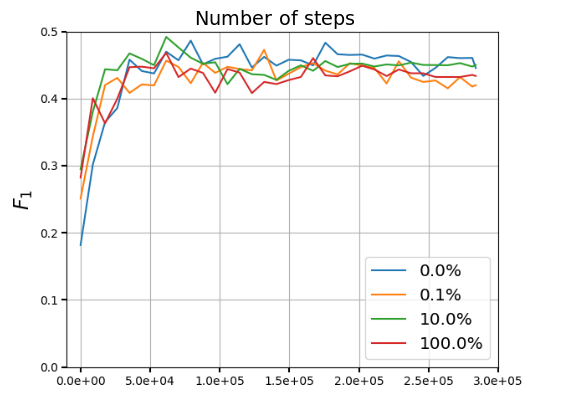}\label{fig:finetune_tass_bert}}
    \caption{Training curves ($F_1$-score) for different amounts of training samples. Figures (a), (b) and (c) shows the fine-tuning results on the validation split for the biLSTM model and Figures (d), (e) and (f) for BERT. (a) and (d) are the validation scores on MeLiSA, (b) and (e) are the scores on the MuchoCine validation split and (c) and (f) the scores on the TASS validation.}
    \label{fig:finetunning}
\end{figure*}

\subsection{Classification Models}

As mentioned in Section \ref{sec:intro}, we explored the cross-domain analysis using two different neural-based classification models which are shown in Fig. \ref{fig:classification_models}.

\begin{itemize}
    \item \textbf{BiLSTM classifier}. This architecture is based on a bidirectional recurrent neural network with a LSTM unit activation and it is illustrated in Figure \ref{fig:blstm_classifier}. In this model, each word $w_t$ of the input sequence $w_1,\ldots,w_T$ is represented as a continuous vector $\mathbf{x}_t$ trough an embedding layer at the beginning of the network. This vector sequence is then forwarded to two different LSTM networks~\cite{lstm} (one forward and one backward). The outputs at the last step of these layers are then concatenated and forwarded to the linear output layer, which gives the probabilities of each class through a Softmax activation function. 
    \item \textbf{BERT classifier}. Since the emergence of the Transformer architecture~\cite{transformer}, a series of models based on self-attention mechanisms have been proposed to pre-train a language model. One of this models is the Bidirectional Encoder Representations from Transformers (BERT), illustrated in Figure \ref{fig:bert_classifier}, which consists of 12 identical transformer encoder layers. These layers contains a multi-head self-attention layer~\cite{transformer} at the input followed by a linear layer (feed forward) with some residual connections and layer normalization~\cite{layernorm} in between. As in the LSTM classifier, every word at the input of the network is represented as a continuous vector, but additionally some special tokens are added to the sentence. In particular, the \texttt{[CLS]} token is included at the start of every sentence and it is used to extract features of the entire sequence. That is, at the output of the encoder a sequence of the same length as the input is obtained but only the first vector is used as an input of the output layer, which returns the class probability.
\end{itemize}

The key difference in our analysis of these two models is that the biLSTM network was trained from scratch, whereas the BERT classifier used was pre-trained on a Language Modeling task. Specifically, the pre-trained model called BETO~\cite{beto}, which is trained on a 3 billion words corpus called Spanish Unnanotated Corpora\footnote{https://github.com/josecannete/spanish-corpora} (SUC) was used. %The SUC database is a collection of unnanotated documents, most of them extracted from the Spanish portions of the Open Parallel Corpus (OPUS) subcorpora. The OPUS project is intended to provide the community with a publicly available parallel corpus of free online data. 
Documents included in the SUC contains all the data from Spanish Wikipedia available at the time the corpus was released and all of the sources of the OPUS Project~\cite{opus} that had text in Spanish. This sources
includes United Nations and Government journals, TED Talks, Subtitles, News Stories and more. However, none of these sources are review-like documents. The total size of the corpora gathered was comparable with the corpora used in the original BERT.
% @misc{cardelino,
%     title = "Spanish Billion Words Corpus and Embeddings",
%     author = "Cristian Cardellino",
%     year = "2016",
%     month = "March",
%     URL = "https: //crscardellino.github.io/SBWCE/"
% }

\begin{table*}
    \centering
    \caption{Test results ($F_1$-score) in the fine-tuning configuration.}
    \label{tab:melisa_finetunning}
    \begin{tabular}{r|cc|cc|cc}
        & \multicolumn{2}{c|}{Amazon} & \multicolumn{2}{c|}{TASS} & \multicolumn{2}{c}{MuchoCine} \\
        (\%) & biLSTM & BERT & biLSTM & BERT & biLSTM & BERT \\\hline
        0.0 & 0.5594 & 0.5860 & 0.3470 & 0.3850 & 0.2441 & 0.2758 \\
        0.1 & 0.5643(+0.88 \%) & 0.5860(+0.00 \%) & 0.3557(+2.51 \%) & 0.4104(+6.60 \%) & 0.1868(-23.47 \%) & 0.2834(+2.76 \%) \\
        10.0 & 0.5654(+1.07 \%) & 0.5865(+0.09 \%) & 0.3761(+8.39 \%) & 0.3974(+3.22 \%) & 0.2255(-7.62 \%) & 0.3210(+16.39 \%) \\
        100.0 & 0.5662(+1.22 \%) & 0.5845(-0.26 \%) & 0.3503(+0.96 \%) & 0.4045(+5.06 \%) & 0.2616(+7.17 \%) & 0.3125(+13.31 \%) \\
    \end{tabular}
\end{table*}

\subsection{Experimental Set-Up}

The above mentioned models were used to perform CDSC using the fine-tuning and the zero-shot configurations. For the fine-tuning case, the following steps were applied:
\begin{enumerate}
    \item The model was trained using the train split of the source dataset (MeLiSA, for instance) and hyperparameter search was done with its corresponding validation portion.
    \item Once trained, the model was trained again using the training portion of the target dataset (MuchoCine, for instance), and a new hyperparameter search was done with the target validation split.
    \item Once retrained, the model was evaluated on the test split of the target dataset.
\end{enumerate}

The only difference between both configurations is that step 2 is omitted for the zero-shot learning. This means that evaluation on the target dataset was done without using any training sample of that dataset. As a consequence, zero-shot learning is usually more challenging than fine-tuning and tends to show lower performance. However, it can provide a better idea of the model's generalization capability.

In order to test if CDSC can be achieved from product domains to more general domains like movie reviews or tweets, we used our MeLiSA dataset as source domain and MuchoCine and TASS as target domains. We also used the Amazon dataset as target domain to keep track of the model's learning capability, although this domain is, in principle, be very similar to the source domain. 

Experiments were carried on in \texttt{Python}, and the \texttt{Pytorch} module was used to implement the model training algorithm. We also used the \texttt{Huggingface Transformers} library to load the Spanish BERT pre-trained parameters and a NVIDA GTX 1080 GPU to reduce time computation. We followed \cite{random_grid_hyperparms} to perform random grid sample hyperparameter search on the biLSTM model. The best biLSTM model found consisted in a two-layer LSTM cell with hidden dimension of 50 and an embedding matrix of dimension $60,\!000\times 300$. Dropout was used as a regularization technique with a probability of $0.1$ and Adam Optimization with a batch size of 16 and a learning rate of 1e-3 was found to give the best validation results. For the pre-trained BERT model, layer dimensions are fixed in advance (12 layers with inner dimension of 768). We also used Adam Optimization to train this model and fixed the learning rate and the batch size to 5e-5 and 16 respectively, as suggested in \cite{beto}.

\section{Results and Discussion}\label{sec:results_discuss}

To provide a better understanding of the cross-domain analysis we evaluated both models as a function of the number of samples used in the training stage. I.e., we kept track of the model's performance when trained with a specific percentage of MeLiSA samples. To measure the classification performance, we used the $F_1$-score, which takes into account possible class unbalances.

\subsection{Fine-tuning}

Figures \ref{fig:melisa_pretrain} and \ref{fig:melisa_pretrain_bert} show, for the biLSTM and BERT models respectively, the $F_1$-score on the MeLiSA validation split as a function of the training steps. This score is shown for trainings with different amount of MeLiSA samples. This first experiment was developed not only to pre-train a model in the source domain, but also to check if adding more samples improves performance in the same domain. It can be seen that this is true for validation and that BERT outperforms the biLSTM model in this task.

Once the model was pre-trained, we analyzed if it could be used to improve performance in other domains. That is, we compared if a model pre-trained on the MeLiSA training set and fine-tuned on the training set of the target dataset performs better than the same model trained with the target training set only. Figures \ref{fig:finetune_cine} and \ref{fig:finetune_tass} show the training curves of the biLSTM model for the MuchoCine and TASS datasets, while Fig. \ref{fig:finetune_cine_bert} and \ref{fig:finetune_tass_bert} show the same curves for BERT. All of these plots represents the $F_1$-score as a function of training steps for different amounts of MeLiSA samples. It should be noticed that the blue curve in these four plots represents the performance of the model without any addition of MeLiSA samples (0.0\%). 

In addition to these plots, we also evaluate the models on the test set of all target datasets. Table \ref{tab:melisa_finetunning} shows the $F_1$-score for the model trained only on the target training set and the relative increment (\%) of this score when more samples of MeLiSA are added to the training. Except for a few specific cases, this table shows that the addition of training samples from MeLiSA improves the $F_1$-score when the model is fine-tuned to other domains.

It is interesting to notice some common patterns between plots in Fig. \ref{fig:finetunning} and Table \ref{tab:melisa_finetunning}. In the first place, there is a clear superiority of the BERT model above the biLSTM in terms of $F_1$-score, which can be seen in all the cases where both models were trained with the same amount of MeLiSA samples. However, it should be take into account that BERT has been pre-trained with significantly more data than the biLSTM model, so a further analysis is needed to make a fair comparison between both models. 

Secondly, we observe that adding more samples to the biLSTM model not always improves its performance. This can be anticipated from Fig. \ref{fig:finetune_cine} (where the target domain are movie reviews from the MuchoCine dataset) and confirmed in Table \ref{tab:melisa_finetunning}. The plot shows that the blue line (0\% of additional samples) is above the yellow one (0.1\% of additional samples) by a significant margin and the table shows that a biLSTM model trained without any additional samples ($F_1=0.2441$ for 0\% of additional MeLiSA samples) performs better than a model trained with some additional samples from the product reviews domain ($F_1=0.1868$ for 0.1\% of additional MeLiSA samples and $F_1=0.2255$ for 10\%). Table \ref{tab:melisa_finetunning} shows a similar behaviour for the tweets domain (TASS column), where the addition of MeLiSA samples does not increase the performance in a proportional amount. 

In summary, Table \ref{tab:melisa_finetunning} suggests that the overall result of using additional training samples from the product reviews domain to pre-train a model in the tweets or movie reviews domains is better in terms of $F_1$-score performance. However, the specific number of samples needed to considerably improve performance may depend on the specific domain and the model used.

\begin{figure*}
    \centering
    \subfloat[]{\includegraphics[height=.35\textwidth,width=.35\textwidth]{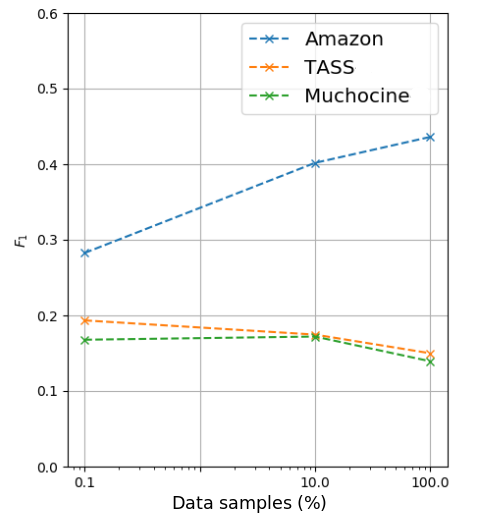}\label{fig:lstm_zero_shot}}
    \hfil
    \subfloat[]{\includegraphics[height=.35\textwidth,width=.35\textwidth]{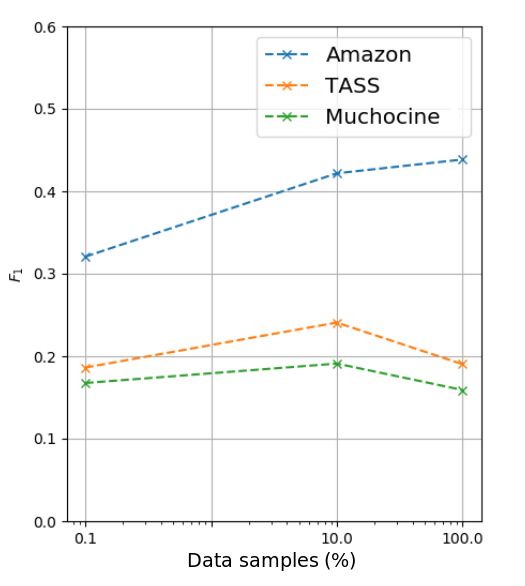}\label{fig:bert_zero_shot}}
    \caption{Zero shot results on test split. Plots show the $F_1$-score as a function of the percentage of training samples for the (a) biLSTM and (b) BERT models.}
    \label{fig:zero_shot}
\end{figure*}

\subsection{Zero-shot}

We also studied each model's performance on the zero-shot CDSC task. In this case, models pre-trained in the context of product reviews were directly tested on the test portion of the Amazon, TASS and MuchoCine datasets. Results are shown in Fig. \ref{fig:lstm_zero_shot} and \ref{fig:bert_zero_shot} for the biLSTM and BERT model, respectively.

There are two main things to point out about this experiment. The first one is that zero-shot classification seems to be significantly better (at least in terms of $F_1$-score) when knowledge is transfered to the same domain. It is interesting to notice, however, that both models perform similar when all samples are used in the pre-training step but BERT has a better performance when less data is used at that step. This could have some relation with the Amazon results shown in Table \ref{tab:melisa_finetunning} from the fine-tuning Section, where the relative improvement of the biLSTM model seems also to be higher than the one of BERT. This may suggest that adding more samples to a model could result in a better relative improvement (in terms of $F_1$-score) when the model is less pre-trained, even if the samples are taken from a different (though from the same domain) dataset.

Another thing to notice is that CDSC is very challenging when there is no fine-tuning adaptation to the target domain, which can be shown in the TASS and MuchoCine curves of both models. It seems to be, however, a slightly improvement of the performance when the percentage of data samples is increased from 0.1\% to 10.0\% for the BERT model. It is not clear from our study if this could mean something relevant but it could provide some additional evidence that a better pre-trained model could take advantage of additional data from other domain, which is also suggested by Table \ref{tab:melisa_finetunning}'s results.

\section{Conclusions}\label{sec:conclusions}

In this work we performed a preliminary study on the Spanish Cross-Domain Sentiment Classification task, which is an important topic in the field of Natural Language Processing but that has not been received enough attention for the Spanish language. Specifically, we were interested in analyzing the knowledge transfer from the product reviews domain to the tweets and movie reviews domains. To address this problem, we created a dataset called MeLiSA (MercadoLibre for Sentiment Analysis) which consists in product reviews from seven Latin American countries and used it to train two different neural-based classification models. We then tested them on three other datasets, which included the Spanish portion of the Multilingual Amazon Reviews Corpus (product reviews domain), the MuchoCine corpus (movie reviews domain) and the 2012 release of the TASS corpus (general content tweets domain).

We considered two models in this work. The first one, was a model  pre-trained on a Language Modeling task transformer encoder (BERT) and the second one a bidirectional recurrent model trained from scratch (biLSTM). In order to perform a complete cross-domain analysis, both models were tested in the fine-tuning and the zero-shot configurations. Results suggest that cross-domain classification can be improved by fine-tuning the model to the specific target domain, but the number of samples needed to achieve optimal results may depend on the target domain and the model used. On the other hand, zero-shot classification seems to perform well on the same domain but very poorly on other domains when trained with product reviews. However, from the performed experiments, we hypothesize that bigger pre-trained models could take more advantage from source domain samples than not pre-trained models.

A possible continuation of this work may include further testing of this last hypothesis by including other pre-trained and not pre-trained models. In particular, it would be useful to test the cross-domain classification on a model trained from scratch and the same model pre-trained on a Language Modeling task. %We also think that it would be a good contribution to the community to make MeLiSA public, but a series of ethical issues should be considered first. \textcolor{red}{Leo: Quizas la ultima oracion la sacaria}

\bibliographystyle{IEEEtran}
\bibliography{IEEEabrv,references.bib}

\end{document}